# Evaluation of a 1-DOF Hand Exoskeleton for Neuromuscular Rehabilitation

Xinalian Zhou, Ashley Mont, and Sergei Adamovich

New Jersey Institute of Technology, Newark NJ 07102, USA
alexzhou@njit.edu

**Abstract.** A low-cost 1-DOF hand exoskeleton for neuromuscular rehabilitation has been designed and assembled. It consists of a base equipped with a servo motor, an index finger part, and a thumb part, connected through three gears. The index part has a tri-axial load cell and an attached ring to measure the finger force. An admittance control scheme was designed to provide intuitive control and positive force amplification to assist the user's finger movement. To evaluate the effects of different control parameters on neuromuscular response of the fingers, we created an integrated exoskeleton-hand musculoskeletal model to virtually simulate and optimize the control loop. The exoskeleton is controlled by a proportional derivative controller that computes the motor torque to follow a desired joint angle of the index part, which is obtained from inverse kinematics of a virtual end-effector mass driven by the finger force. We conducted parametric simulations of the exoskeleton in action, driven by the user's closing and opening finger motion, with different proportional gains, end-effector masses, and other coefficients. We compared the interaction forces between the index finger and the ring in both passive and active modes. The best performing assistive controller can reduce the force from around 1.45N (in passive mode) to only around 0.52N, more than 64% of reduction. As a result, the muscle activations of the flexors and extensors were reduced significantly. We also noted the admittance control scheme is versatile and can also provide resistance (e.g. for strength training) by simply increasing the virtual end-effect mass.

## 1   Introduction

Stroke, one of the leading causes of adult disability, affects approximately 800,000 individuals each year in the United States [1]. Nearly 80% of stroke survivors suffer from hemiparesis of the upper arm and thus impaired hand function, which is integral to most activities of daily living. It is well established that highly repetitive training can aid in the recovery of motor function of the hand however this can be labor intensive for the providing physical therapist in addition to the cost. In the past decade, more robotic hand rehabilitation devices have been introduced to help patients recover hand function through assistance during repetitive training of the hand [2-4].

In a comprehensive review by Heo et al. [2], hand exoskeleton technologies for rehabilitation and assistive engineering, from basic hand biomechanics, neurophysiology, sensors and actuators, physical human-robot interactions and ergonomics, are summarized. Different types of actuators and control schemes have been used for hand exoskeletons. In some control schemes, the robotic device will move the user passively through a preprogrammed trajectory for continuous passive movement (CPM) therapy. These devices can be beneficial for severely impaired individuals who may not have the ability to generate the forces required for specific finger or hand movement or for individuals who have abnormal muscle synergies preventing continuous movement. A few devices such as the Kinetic Maestra and Vector 1 are commercially available devices that are used for CPM [5, 6]. These devices allow for passive movement through the range of motion for individual fingers. However, as there is no active participation by the user, this device on its own may not promote neurorehabilitation. These devices can be combined with other simulations or control schemes that require active participation by the user. One commercially available hand exoskeleton that has been used extensively by our lab to provide haptics to virtual simulations is the CyberGrasp [7]. The CyberGrasp is a cable driven exoskeleton that weighs 450 grams and can provide up to 12 N of force on each finger and can be used to provide assistance for extension of the user's fingers. In one study, this was used in combination with a virtual reality simulation to train finger individuation as the user played a virtual piano [8]. The CyberGrasp was used to resist finger flexion of the inactive fingers, promoting movement of the active independent finger. Similarly, the eXtension Glove (X-Glove) was developed to be used for cyclical stretching in addition to active movement training [9-11]. This cable driven design is actuated using linear servos allowing for individual finger movement in both extension and flexion. In addition to this, each cable is integrated with a tension sensor which allows the force of each digit to be monitored. This device has two modes that can be used for rehabilitation, the first mode cyclically extends and flexes the fingers. The second mode is an active training mode in which the glove provides constant extension assistance so that the user can complete flexion tasks as long as they overcome the force required to keep the finger extended. In a further attempt to integrate user control with the exoskeleton, an external input from the user such as force or electromyography has been incorporated into some designs such as the Helping Hand [12]. This soft robotic device allows for active assistance for each finger individually, in addition to the ability to follow control states triggered by EMG.

In this paper, we introduce a low cost 1-DOF hand exoskeleton for neuromuscular rehabilitation of individual fingers. This exoskeleton consists of a base equipped with a servo motor, an index finger component and a thumb component connected with gears. The exoskeleton's control system was designed to generate suitable actuation torques based on the interaction force between the user's finger and the exoskeleton's index component. The goal of this study is to model the exoskeleton interacting with a neuromuscular hand model in order to evaluate the effectiveness of an intuitive admit-tance control algorithm on providing different levels of assistance or resistance during hand rehabilitation.

## 2 Methods

### 2.1 The 1-DOF Exoskeleton and Hand Model

This exoskeleton consists of a base stationed with a servo motor (Dynamixel XM430), an index finger part and a thumb part, which are connected through 3 gears of equal sizes as shown in Fig. 1. The motor drives the top gear which in turn rotates the gear attached to the index part and then the gear attached the thumb part. The index and thumb parts both have rings for the fingers, and an OptoForce tri-axial load cell or force sensor (OnRobot, Denmark) is attached to the index ring. All parts are 3D printed with a carbon fiber reinforced nylon material called Onyx (Markforged, USA). The total weight of this exoskeleton is 0.158kg and the mass and inertia properties of its components, which were either measured or computed based on material and part geometry, are listed in Table 1.

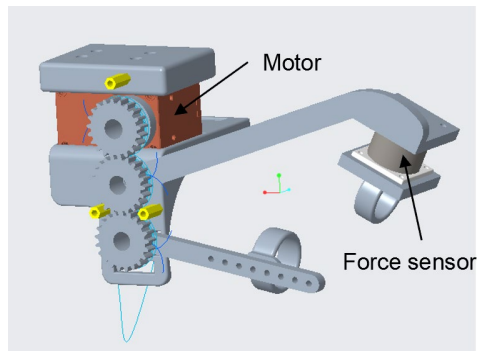

**Fig. 1.** The design of the 1-DOF hand exoskeleton.

Table 1. Mass and moment of inertia (MOI) properties of exoskeleton components. x: fore-aft; y: vertical; z: lateral. (MOI unit: $kg \cdot cm^2$)

| Exo-Part | Mass (kg) | $I_{xx}$ | $I_{yy}$ | $I_{zz}$ |
|---|---|---|---|---|
| Base | 1.22e-1 | 6.62e-1 | 8.39e-1 | 6.94e-1 |
| Index | 2.95e-2 | 6.11e-2 | 5.56e-1 | 5.58e-1 |
| Thumb | 6.40e-3 | 9.20e-3 | 6.51e-2 | 6.43e-2 |

The exoskeleton was modeled as an articulated rigid body system with just one true rotational DOF at the motor (top) gear and its base fixed on the human hand. To model the interaction between the human hand and the exoskeleton, we used a generic hand musculoskeletal model adapted from the one developed by Lee et al. [13]. The adapted hand model, shown in Fig. 2, has six finger muscles including: extensor digitorum communis (EDC), extensor indicis (EI), flexor digitorum superficialis (FDS), flexor digitorum profundus (FDP), bipennate first dorsal interosseous (FDI) on the radial side and the ulnar side. In the original model by Lee et al., only the radial side of the FDI, which connects the radial side of the second metacarpal to the radial side of the base of the second proximal phalanx (index finger), was modeled. Here we added the opposite (ulnar) side of the FDI that connects the proximal half of the ulnar side of the first metacarpal (thumb) to the index finger as we believe this branch could be important for finger closing and opening motion. Among these muscles, EDC and EI are extensors and FDS, FDP and FDI are flexors. To assemble with the hand musculoskeletal model (Fig. 2), the exoskeleton base part was fixed on the back of the hand through a fixed joint while the ring on the index part was put on the index finger, and similarly for the ring on the thumb part.

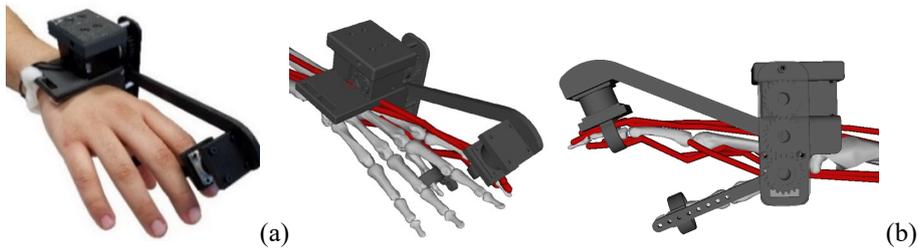

(a)          (b)

**Fig. 2.** (a) The hand exoskeleton device put on a human hand. (b) The assembled model of the hand musculoskeletal system and the exoskeleton.

### 2.2 Admittance Controlled Exoskeleton

To help users with muscle weakness during training, we designed an admittance control paradigm, shown in Fig. 3, to provide intuitive control and positive force amplification to assist the user's finger movement. This admittance control transfers the force from the load cell into the motion of a virtual end-effector mass in the task space. The admittance control framework integrates the motion of this end-effect mass initially placed at the center of the index ring and moved by the force applied by the index finger. The desired angle of the motor ($\theta_d$) is obtained from an inverse kinematics (IK) computation based on the desired end-effect position, and the motor is then commanded to achieve this position. In hardware implementation, the end-effect force is measured by a force/torque sensor and the desired motor angle $\theta_d$ can be achieved by either torque based control or position based control of the Dynamixel motor. In this work, we evaluated this admittance control loop with computer simulations. The interaction forces between the finger and the ring were modeled and will be described shortly. And a proportional derivative (PD) controller was used to prescribe a desired joint torque ($\tau$) to the motor:

$$\tau = k_p(\theta_d - \theta) - k_d \dot{\theta} \qquad (1)$$

where $k_p$ and $k_d$ are tunable parameters, $\theta$ and $\dot{\theta}$ are the current angle and angular velocity of the motor, respectively.

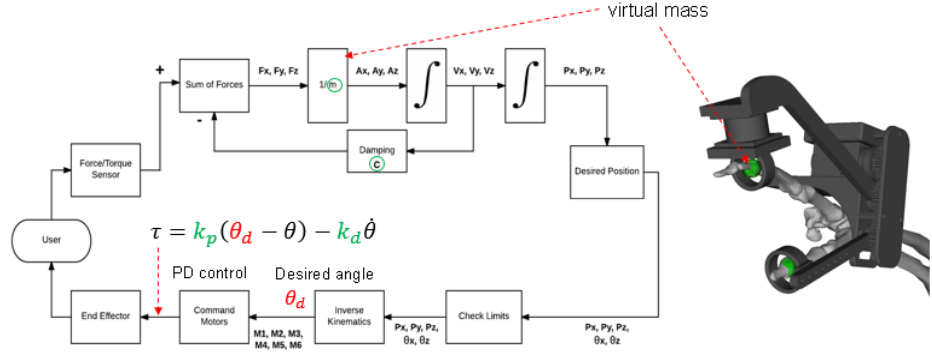

**Fig. 3.** The designed admittance control framework. The parameters in green or circled in green are tunable control parameters.

To model the interaction forces between the finger and ring, we introduced a tri-directional spring-damper force element that mimics the contact between them. The tri-directional force element was introduced at the center of the ring and can predict directional differences in force responses due to the relative movement of the finger and the ring. The force element computes three (XYZ) directional distances between a point on the exo-part and its counterpart on the finger and generates (either positive or negative) forces along these directions during their relative movement. At the initial assembly, these two points are coincident to each other and generate zero force ($x_0 = y_0 = z_0 = 0$). The forces generated by a force element were modeled by linear damped springs:

$$\begin{cases} f_x = k_x(x - x_0) + c_x \dot{x} \\ f_y = k_y(y - y_0) + c_y \dot{y} \\ f_z = k_z(z - z_0) + c_z \dot{z} \end{cases} \quad (2)$$

The stiffness and damping constants of the directional force element are listed in Table 2. The stiffness in the lateral direction (YZ) is assumed to be 20 times of that in the X direction to mimic the behavior of harder resistance in the lateral directions and softer resistance in the sliding directions.

Table 2. Stiffness and damping of the direction spring between finger and ring.

| Directional springs | Stiffness (N/m) | | | Damping (Ns/m) | | |
| --- | --- | --- | --- | --- | --- | --- |
| | $k_x$ | $k_y$ | $k_z$ | $c_x$ | $c_y$ | $c_z$ |
| Finger-Ring | 500 | 10000 | 10000 | 80 | 200 | 200 |

### 2.3 Simulation Methods

With the developed model and simulated controlled framework, we can perform computer simulations to study the co-operation between fingers and the exoskeleton. In our simulations, we assumed the index finger moves to actively track a given closing and opening motion. The thumb is passive and moved by the thumb ring and the ulnar side of the FDI muscle. The motion of the exoskeleton is driven the by the motor torque and the forces between the fingers and the rings. At the passive mode, the active motor torque was set to zero and a small velocity dependent damping was applied.

All simulations in this study were performed with our in-house musculoskeletal simulation code, CoBi-Dyn. A hybrid inverse dynamics (ID) and forward dynamics (FD) simulation framework similar to the one presented in [14] was employed. The human finger joints were classified as ID joints such that their motions can be prescribed to track an input motion. The exoskeleton joints were classified as FD joints such that their motions were driven by the actuation forces and finger-ring interaction forces. At each time step, the hybrid dynamics framework predicted joint torques for all finger joints and accelerations for all exoskeleton joints. The predicted finger joint torques were the target or desired torques that ideally shall be generated from muscles spanning these joints. To compute muscle forces, one of the goals was to find an appropriate muscle force combination that contributed to generate the desired joint torques as closely as possible. Due to the redundancy of the muscles, there could be many such combinations and thus muscle forces were determined by solving an optimization problem. The final objective of this optimization problem was to minimize an objective function, defined as

$$\sum_{i=1}^{n} \left(\frac{f_i}{f_i^{max}}\right)^p + w\boldsymbol{C}^T\boldsymbol{C} \qquad (3)$$

where $f_i$ was the force of the ith muscle, $f_i^{max}$ was the maximum attainable muscle force at its current state, $\boldsymbol{C}$ was the difference vector between the desired joint moments and the moments generated by spanning muscles ($\boldsymbol{C}$ is often called the residual torque); and $w$ was a weighting or penalty factor. And $\frac{f_i}{f_i^{max}}$ can be considered as the muscle activation or effort for simplicity. For all our simulations, $p = 2$ and $w = 100$ were utilized.

### 2.4 Experimental Data Collection

For calibration of model parameters and validation of simulations, we collected experimental data of finger motion with the exoskeleton under passive mode (no torque control). MATLAB 2018a was used to control the gripper as well as obtain data from the motor and tri-axial load cell. The Matlab script was reading the force between the index finger and the exoskeleton measured by the OptoForce three-degree-of-freedom transducer (OnRobot, Denmark). The force, as well as the rotation angle of the motor, were read in a loop at 155 Hz. The EMG data were collected using the Trigno wireless system with Quattro electrodes (Delsys, USA) at 2000 Hz, and synchronized in time with the motor and tri-axial load cell collection in MATLAB using an external trigger. EMG electrodes were placed on the FDI, EI, EDC, and FDS muscles. During the experiments, subjects were seated comfortably in a chair with armrests so that he or she could rest his or her hand in between sessions. Subjects were asked to wear the exoskeleton on their right hand, with their index finger and thumb placed comfortably in the respective rings and the gear axes approximately aligned with the gravity direc-tion. To calibrate the motor positions, the subjects were asked to fully extend their fingers to an open position and then to flex their fingers to position where their thumb met their index finger. These values were recorded and set as limits so that the motor would not exceed these limits as an additional safety precaution in addition to the mechanical stops. The force sensor was then calibrated by collecting 1000 samples and taking the average of these samples as a bias reading. This value was subtracted from the force readings during the data collection to reduce the inherent bias of the ring attachment. For each session, the subject was asked to complete 15 to 20 extension-flexion cycles in synchronization with a metronome set to four of the following speeds: 40 bps, 50 bps, 70 bps, 100 bps or 150 bps. Under the no torque condition, power was provided to the motor but no torque was applied, allowing this to act as a passive exoskeleton where the user had full control over movement. For each session, EMG, force, and motor position were collected. The motor position is determined by the smart servo rotary motor, and through the provided conversion factors the motor joint angle can be calculated in radians. For the tri-axial load cell, we mainly looked at the perpendicular force (Z force), which is aligned with the direction of movement for the index finger.

## 3  Results and Discussion

### 3.1  Experimental Data of Passive Exoskeleton

Using MATLAB 2018a, a custom script was used to process and analyze all session data from one subject. Each session consisted of approximately 15 to 20 open and close cycles and in order to further analyze the data, this had to be divided into individual cycles using peak detection of the motor position. Motor position was resampled to 350 Hz and filtered with a $4^{th}$ order lowpass Butterworth filter with a cutoff of 8Hz. The start of the cycle was considered to be the instant when the motor position was at its extreme and the index finger was fully extended. The entire cycle included the flexion to extension movements and the velocity was derived from the position values. The force data were also resampled to 350 Hz and filtered with a $4^{th}$ order lowpass Butterworth filter with a cutoff of 10Hz. The position, velocity, and force data were then split into individual time synchronized cycles, the average values were obtained. The EMG data were filtered with a $4^{th}$ order highpass Butterworth filter with a cutoff of 20Hz and a $4^{th}$ order lowpass Butterworth filter with a cutoff of 500Hz. The data were then resampled to 350 Hz, rectified, and split into individual cycles that align in time with the position, velocity, and force data. Further, for each muscle, the root mean square envelope was calculated using a sliding window of 30 samples, with an overlap of 29 samples. As a maximum voluntary contraction was not obtained, the root mean square (RMS) envelope was normalized to the maximum mean value of the FDI muscle.

During a no torque session at 40 bps, on average, the subject was able to move through a rotational motor angle of approximately 16 degrees (Fig. 4) within approximately 1.5 seconds. At the peak of the flexion phase, the user is applying approximately 1.2 N, and at the peak of the extension phase the user is applying .5 N (Fig. 5).

This greater force applied during flexion can be attributed to a slower velocity during extension than during the flexion phase. Although the subject was listening to the metronome, it is possible that their speeds change between flexion and extension due to error. Additionally, it is possible that the friction of the motor is slightly higher in the direction of rotation for flexion requiring the user to apply a larger force. The muscle activity of the FDI, EI, FDS, and EDC muscles can be seen in Fig. 6. As expected, we see an onset of activation of the index flexor muscle, FDI, during finger flexion, and relaxation during the extension phase. Similarly, for the EDC muscle we see an onset during the extension phase of movement, and relaxation during flexion.

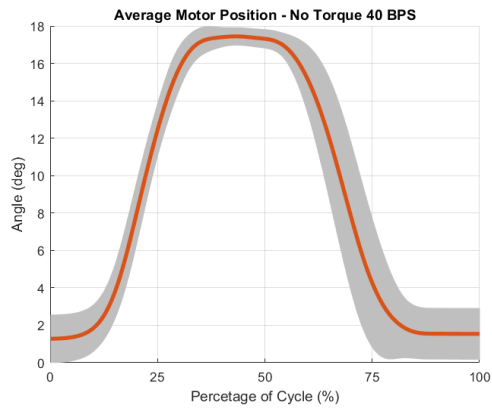

**Fig. 4.** Average motor rotation angle of the exoskeleton during the closing and opening cycles. The grey shaded area indicates the variance.

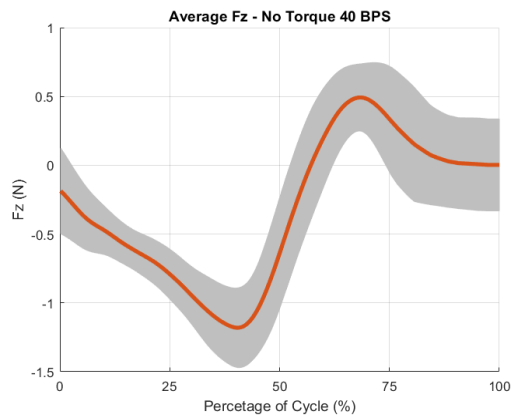

**Fig. 5.** Average measured force from the tri-axial force sensor during the closing and opening cycles. The grey shaded area indicates the variance.

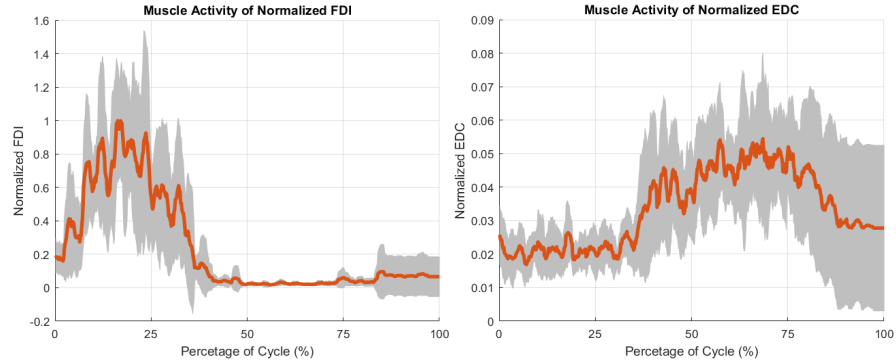

**Fig. 6.** Average RMS muscle EMG (normalized by the maximum RMS value of FDI) during the closing and opening cycles. The grey shaded area indicates the variance.

### 3.2  Simulation Results

We conducted parametric simulations of the exoskeleton in action, driven by the fingers' closing and opening movement and torque control of the motor, with different combinations of the proportional gain and damping of the PD controller, the end-effector mass, and the end-effect damping coefficient. The virtual end-effect mass ($m$) values were selected from 0.01 to 10 $kg$. For simplicity, the unit maybe ignored in the context below. For values smaller than 0.01 (e.g. 0.001), the control tends to be highly oscillatory and has deteriorated performance. The damping coefficient ($c$) is chosen to be 0.01 for all cases presented here. According to our numerical tests, varying the end-effect damping coefficient seems to have non-significant effects on the results. And the PD controller's damping coefficient $k_d$ seems to have a negative effect on the controller's performance and thus only zero damping cases were presented here.

During the simulations, the index finger metacarpophalangeal (MCP) joint tracks a closing and opening motion. The other two joints of the index finger, the proximal interphalangeal (PIP) and distal interphalangeal (DIP) joints, were assumed to be stationary (zero angle). All three joints are ID joints, which means their motions are given and torques are computed from ID. Fig. 7 shows the angle and angular velocity of the input motion for the MCP joint, which symmetrically increases from 0 to 25 degrees during closing and decreases from 25 to 0 degree during opening, with a duration of 1.5 seconds. The angle of the motor, which was computed from the FD, is also shown in Fig. 7. As it can be seen, it is slightly less than 14 degrees, which is close to the experimental measurement. The difference in angles between the finger and motor was caused by their geometry differences and joint locations.

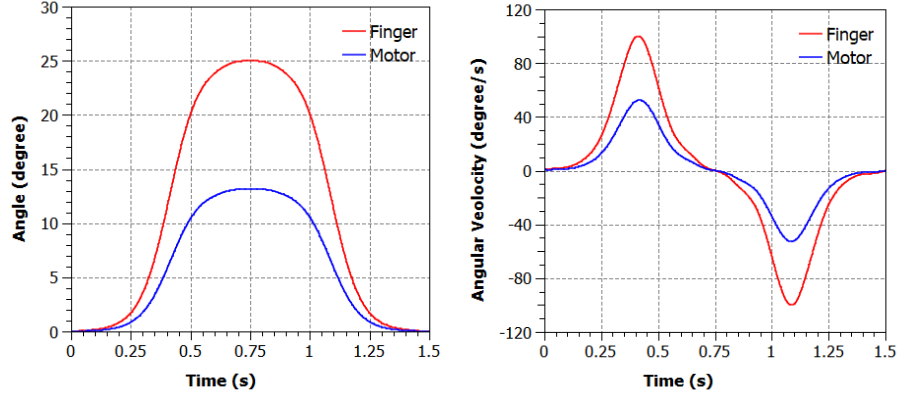

**Fig. 7.** Angles and angular velocities of the index finger and the motor.

During the tracking of the input motion, the finger joint torques were computed and subsequently muscle force and activation were predicted from the optimization routine. The interaction force between the finger and the ring is the main factor that affects the torques and thus muscle force prediction. In Fig. 8, interaction forces between the index finger and the ring for different control modes and parameters are presented. For the passive mode, the interaction force is slightly less than 1.45N, within the variance of our experimental measurement. For end-effect mass $m = 0.1$, different PD proportional gains ($k_p = 0.1, 1, 2$) were tested. Compared to the passive case, increasing $k_p$ tends to increase the control performance as it reduces the maximum interaction force. However, once it exceeds a certain value, e.g. $k_p = 2$, the controller will start to oscillate unstably. Similarly, for end-effector mass $m = 0.01$, a reasonable $k_p$ shall be smaller than 1, beyond which unstable oscillation starts. Comparing these controller parameters, the combination of $m = 0.01, k_p = 1$ produced the smallest maximum interaction force at around 0.52N.

We further increased the end-effector mass values to 1 and 10 and compared the performance of the controllers. In Fig. 9, the interaction forces for the four cases ($m = 0.01, 0.1, 1, 10$) are presented. Clearly, increasing the end-effector mass decreases the assistive performance and even produces a resistive force when $m = 10$. With further increase of the mass parameter, the resistance will keep increasing until it reaches the maximum capacity of the motor. This indicates that by simply adjusting the mass parameters, the user can tune the assistance or resistance level as desired. And the optimal mass parameter that provides the best assistance performance is likely the smallest one before the controller starts to oscillate.

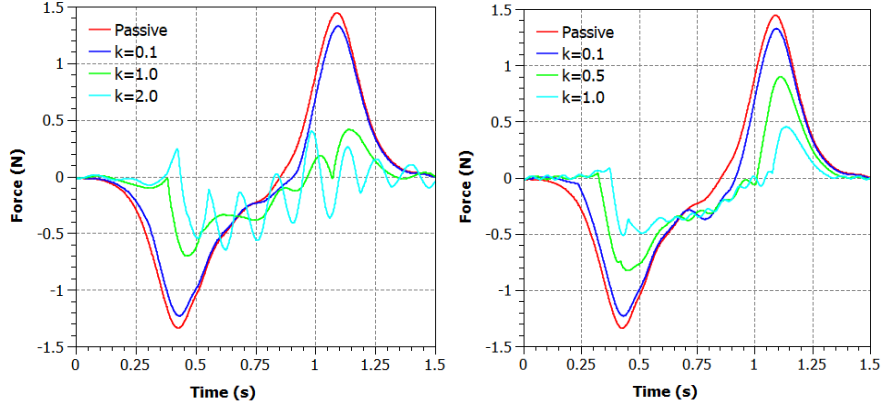

**Fig. 8.** Finger-ring interaction forces for different PD proportional gain ($k_p$ or k in the figure); Left: $m = 0.1$; Right: $m = 0.01$.

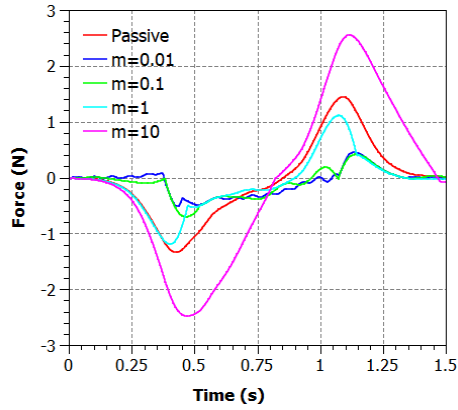

**Fig. 9.** Finger-ring interaction forces for different mass values. $k_p = 1$ for all masses.

In Fig. 10, we compared the active motor torques for several controllers, three of which have $m = 0.01$ and different gains ($k_p = 0.1, 0.5, 1$) and the other with a large $m = 10$ and $k_p = 1$. As $k_p$ increases, the active motor torque increases and provides better assistance to the finger's closing and opening motion, as is evident from the reduction of interaction forces shown in Fig. 8. For the large mass ($m = 10$), the active torque is similar in magnitude as the best assistive torque ($m = 0.01, k = 1$) but with a negative sign (reversed direction), indicating resistance instead of assistance. Considering an extreme case of near infinite mass value of the virtual end-effector, its movement is very slow and stays near the initial position and consequently, the IK based admittance controller will try to pull the exoskeleton back from its movement direction to the initial position (i.e. resisting any movement).

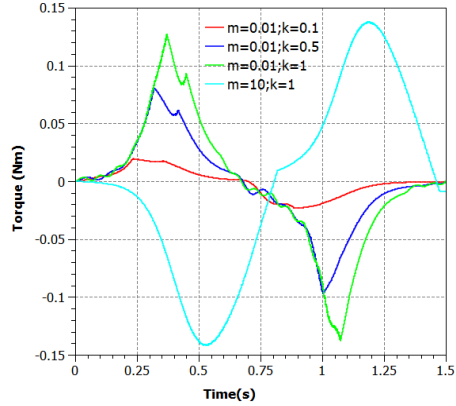

**Fig. 10.** Predicted motor torques for different control parameters. Assistive torques were provided for $m = 0.01$ and a resistive torque was provided for $m = 10$ as is evident from their sign difference.

We also looked at the average muscle activations of the flexors and extensors for these four controllers, as shown in Fig. 11. The flexors are active mostly during the finger closing phase while the extensors are active during the opening phase. In Fig. 12, snapshots of the hand-exoskeleton in motion are shown, with muscle activation rendered in color. For the three assistive controllers with $m = 0.01$, muscle activations of both muscle groups were reduced due to the decreased finger-ring interaction force. For the resistive controller ($m = 10$), the muscle activations for both groups have increased significantly when compared to the passive case.

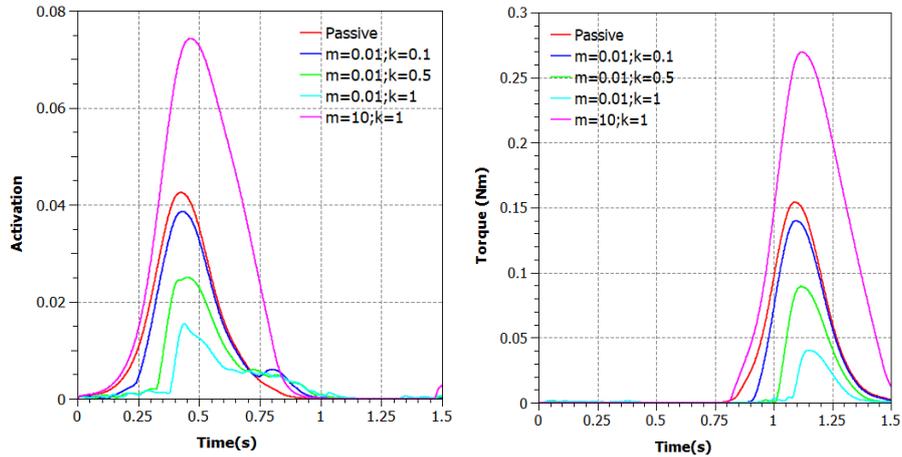

**Fig. 11.** Predicted average muscle activations for different control modes and parameters. Left: flexors; right: extensors.

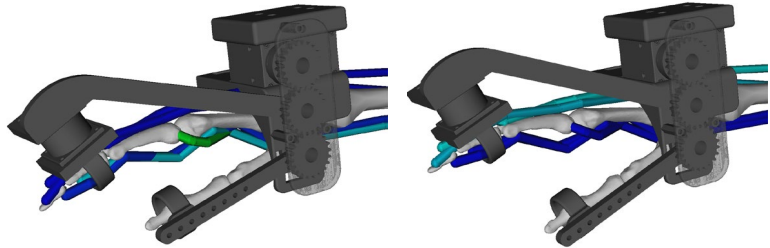

**Fig. 12.** Snapshots of hand-exoskeleton motion and muscle activation (rendered in color) during finger closing (left) and opening (right) movement.

## 4 Conclusion

Through the design and modeling of a 1-DOF hand exoskeleton and its interaction with a hand musculoskeletal model, we were able to evaluate the effectiveness of an admittance control method. The results demonstrated that the assistance provided by the motor reduces muscle activation significantly as a result of reduced interaction forces. Under the current admittance control, increasing the PD controller's proportional gain $k_p$ often results in better assistive performance until it produces overshoot oscillation. And decreasing the virtual mass seems to achieve better assistance performance as well until the occurrence of unstable oscillation. Resistance can also be achieved with the admittance control, by simply increasing the value of the virtual end-effector mass beyond a certain value. In conclusion, modeling can help to predict the feasibility of the admittance control framework, guide the tuning of control parameters, and evaluate the exoskeleton's effectiveness for hand rehabilitation. We are currently implementing the present admittance controller in hardware and conducting tests to calibrate model parameters and validate the simulation predictions. We hope that, with the developed models and additional parametric simulations, it will enable us to fine tune control parameters, explore the design space, and devise novel or optimal control schemes for implantation in hardware.

**Acknowledgment:** This work was supported by NIDILRR (Rehabilitation Engineering Research Center Grant #90RE5021) and by NIH grant R01HD58301.